\documentclass[letterpaper]{article} 
\usepackage{aaai25}  
\usepackage{times}  
\usepackage{helvet}  
\usepackage{courier}  
\usepackage[hyphens]{url}  
\usepackage{graphicx} 
\usepackage{natbib}  
\usepackage{caption} 
\usepackage{newtxmath}
\usepackage{multirow}
\usepackage{threeparttable}
\usepackage{makecell}
\usepackage{algorithm}
\usepackage{algorithmic}

\usepackage{pdfpages}

\usepackage{newfloat}
\usepackage{listings}
\DeclareCaptionStyle{ruled}{labelfont=normalfont,labelsep=colon,strut=off} 
\floatstyle{ruled}
\newfloat{listing}{tb}{lst}{}
\floatname{listing}{Listing}
\title{Self-Attentive Spatio-Temporal Calibration for Precise Intermediate Layer Matching in ANN-to-SNN Distillation}

\author{
	Di Hong\textsuperscript{\rm 1,2,3}, Yueming Wang\textsuperscript{\rm 1,2,3}\thanks{corresponding author.}\\
}
\affiliations{
	\textsuperscript{\rm 1}The College of Computer Science and Technology, Zhejiang University, China\\
	\textsuperscript{\rm 2}Nanhu Brain-computer Interface Institute, Hangzhou, China\\
	\textsuperscript{\rm 3}The State Key Laboratory of Brain-Machine Intelligence, Zhejiang University, China\\
	$\{$hongd, ymingwang$\}$ @zju.edu.cn 
}

\begin{document}
	
\maketitle

\begin{abstract}
	Spiking Neural Networks (SNNs) are promising for low-power computation due to their event-driven mechanism but often suffer from lower accuracy compared to Artificial Neural Networks (ANNs). ANN-to-SNN knowledge distillation can improve SNN performance, but previous methods either focus solely on label information, missing valuable intermediate layer features, or use a layer-wise approach that neglects spatial and temporal semantic inconsistencies, leading to performance degradation.
	To address these limitations, we propose a novel method called \emph{self-attentive spatio-temporal calibration (SASTC)}. SASTC uses self-attention to identify semantically aligned layer pairs between ANN and SNN, both spatially and temporally. This enables the autonomous transfer of relevant semantic information. Extensive experiments show that SASTC outperforms existing methods, effectively solving the mismatching problem.  Superior accuracy results include 95.12\% on CIFAR-10, 79.40\% on CIFAR-100 with 2 time steps, and 68.69\% on ImageNet with 4 time steps for static datasets, and 97.92\% on DVS-Gesture and 83.60\% on DVS-CIFAR10 for neuromorphic datasets.  This marks the first time SNNs have outperformed ANNs on both CIFAR-10 and CIFAR-100, shedding the new light on the potential applications of SNNs.
	\begin{links}
	\link{Code}{https://github.com/ieveresthd/SASTC.git}
	\end{links}
\end{abstract}
	
\section{Introduction}
Spiking Neural Networks (SNNs), considered the third generation of neural networks \cite{maass1997networks}, offer a promising advancement in low-power computing. Unlike artificial neural networks (ANNs), which use continuous-valued activations, SNNs emulate the brain's discrete, spike-based information transmission, making them ideal for event-driven and energy-efficient neuromorphic hardware \cite{TrueNorth}. Two main approaches have emerged for developing supervised deep SNNs: 1) direct training from scratch using surrogate gradients to approximate the discontinuous derivatives of spiking neurons, and 2) ANN-to-SNN conversion, which aligns ANN neuron functions with spiking neurons. Despite progress, a performance gap persists between ANNs and SNNs. To address this, ANN-to-SNN knowledge distillation has been employed to transfer relevant knowledge from ANNs to SNNs.

However, previous distillation methods have either failed to transfer sufficient knowledge or have faced spatial and temporal disparities in semantic information, resulting in degraded performance. This paper introduces a Self-Attentive mechanism to address the semantic mismatch problem by autonomously identifying the most relevant semantic layer patterns across spatial and temporal dimensions and allocating attention based on semantic relevance. The key contributions of this work are summarized as follows:
\begin{enumerate}
	\item We propose a self-attention mechanism to address semantic mismatching during ANN-to-SNN knowledge distillation by autonomously aligning the most relevant layer patterns between ANN and SNN both spatially and temporally.
	\item Through extensive experiments across various settings and prevalent network architectures, our method significantly boosts SNN performance in ANN-to-SNN distillation, surpassing current benchmarks across various datasets, including both static and neuromorphic ones.
	\item Our analysis demonstrates that SASTC successfully achieves semantic matching in ANN-to-SNN distillation, advancing its applications in robust representation.
\end{enumerate}

\section{Related Work}	
\subsection{Direct Training from Scratch}
We briefly summarize some significant achievements in direct training. Lee et al. directly train SNNs in terms of spikes by regarding the membrane potential as the combination of differentiable signals and discontinuous noisy \cite{lee2016training}. Wu et al. use an approximate derivative to construct an iterative LIF neuron model and propose a spatio-temporal backpropagation (STBP) method to train SNNs from scratch \cite{wu2018spatio}. Zheng et al. propose a threshold-dependent batch normalization (tdBN) method for tuning the loss function \cite{zheng2021going}. Rathi et al. propose to optimize the leakage and threshold in the LIF neuron model. Furthermore, many direct training method have been proposed based on designing various surrogate gradients and coding schemes to achieve SNNs with low latency and high performance \cite{wu2019direct}. 

\subsection{ANN-to-SNN Conversion}
P{\'e}rez-Carrasco et al. first map sigmoid neuron model of ANNs into LIF neuron model by utilizing scaling factor, which is determined according to neuron parameters and modified manually \cite{perez2013mapping}. Diehl et al. propose to regulate firing rates of SNNs through weight normalization \cite{diehl2015fast}. Cao et al. adopt only one hyperparamter, which is the firing threshold of spiking neurons, to approximate the rectified linear unit (ReLU) function of ANNs \cite{cao2015spiking}. Based on the great success achieved in previous conversion schemes, many subsequent studies are devoted to minimizing various errors in conversion process \cite{Sengupta2019}. 

\subsection{ANN-to-SNN Distillation}
Typically, SNNs employ the spike frequency of the output layer or the average membrane potential increment as inference indicators. Analogous to ANN distillation, the conventional ANN-to-SNN knowledge distillation minimizes the Kullback-Leibler (KL) divergence between these SNN inference indicators and the predictive class probability distributions of ANNs \cite{LeeKDSNN}. Recent efforts explore the transfer of enriched information from feature maps to enhance performance \cite{hong2023lasnn}.

\section{Self-Attentive Spatio-Temporal Calibration}
\subsection{Notations and Background}
In this section, we provide a concise overview of fundamental concepts and establish necessary notations for subsequent illustration or clarity, the term "teacher model" denotes the ANN model, while the "student model" refers to the SNN model unless explicitly specified. Let $\mathcal{X} = \left\{x_{i}, y_{i}\right\}^{n}_{i}$ represent the training dataset consisting of $n$ instances and $N$ categories, with $x_{i}$ as the input vector and $y_{i}$ as the corresponding target in the form of a one-hot encoding vector. The number of output channels and spatial dimensions represented as $c$, $h$ and $w$, respectively. 
For a mini-batch data of size $b$, the output of each SNN layer $s_{l}$ at time step $t$ is denoted as $f^{t}_{s_{l}} \in \mathbb{R}^{b \times c_{s_{l}} \times h_{s_{l}} \times w_{s_{l}}}$, where the superscript $t$ signifies the index of the current time step, and $T$ represents the total number of time steps. Simultaneously, the output of each ANN (teacher) layer $a_{l}$ is denoted as $f_{a_{l}} \in \mathbb{R}^{b \times c_{a_{l}} \times h_{a_{l}} \times w_{a_{l}}}$. The layer indices $s_{l}$ and $a_{l}$ traverse from $1$ to $s_{L}$ and $a_{L}$, respectively. Notably, $s_{L}$ and $a_{L}$ typically differ due to the intrinsic heterogeneity inherent in the teacher and student models. The output representations at the penultimate layer of the teacher and student models are labeled as $f_{a_{L}}$ and $f_{s_{L}}$. Furthermore, we define the feature pattern $F$ as the set of outputs from intermediate feature layers. $F^{t}_{s}$ represents the feature pattern of the student model at time step $t$, $F^{t}_{s} = \{f^{t}_{s_{l}} ~ | ~ \forall ~ l \in [1, \dots, L]\}$, while $F_{a}$ denotes the feature pattern of the teacher model, $F_{a} = \{f_{a_{l}} ~ | ~ \forall ~ l \in [1, \dots, L]\}$. It is crucial to note that this collection is a permutation rather than a combination. In other words, multiple collections with the same intermediate layers but in different orders signify distinct feature patterns.

Concerning the student model, the outputs of the final layer $f_{end}(\cdot)$ are represented as the averaged membrane potentials over all time steps, $O^{i}_{s} = \frac{1}{T} \sum\limits^{T}_{t=1} f_{end}(f^{t}_{s_{L}}[i]) \in \mathbb{R}^{N}$, where the notation $i$ refers to the $i$-th input instance. We have added this clarification in the revised version. Predicted probabilities are derived through a softmax layer built on these outputs $O^{i}_{s}$, denoted as $P^{i}_{s} = \sigma(O^{i}_{s} / \alpha)$. Similarly, the logits of the teacher model are designated as $O^{i}_{a} = f_{end}(f_{a_{L}} [i]) \in \mathbb{R}^{N}$, and the corresponding predicted probabilities are denoted as $P^{i}_{a} = \sigma(O^{i}_{a} / \alpha)$, commonly referred to as soft targets. In both the student and teacher models, the hyperparameter $\alpha$ is typically set to 1. 

\subsection{Spatio-Temporal Mismatch Problem on Existing ANN-to-SNN Knowledge Distillation}
\begin{figure} [t]
	\centering
	\includegraphics[width=0.473\textwidth]{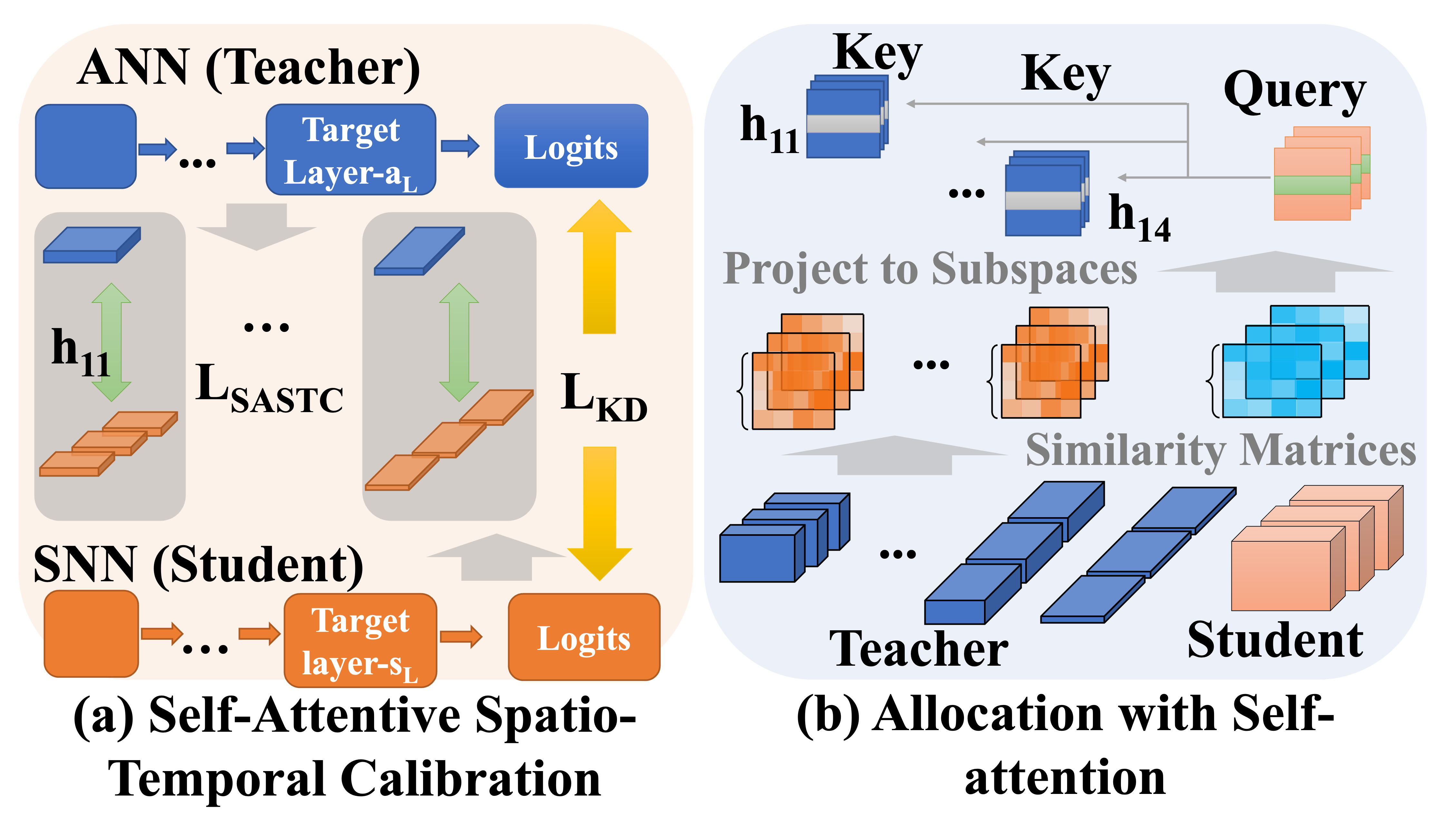}
	\caption{An overview of the proposed Self-Attentive Spatio-Temporal Calibration. }
	\label{fig:overview}
\end{figure}	

Prior studies have assumed that: 1) the distributions of semantic information embedded in ANNs and SNNs are similar (spatially matched), and 2) this distribution within SNNs remains constant across different time steps (temporally matched). We introduce a metric named Spatio-Temporal Mismatch Score (STM score) to assess the extent of semantic disparity between associated ANN-SNN layer pairs over time steps. STM score is computed as the Average Euclidean Distance between the generated similarity matrices of each corresponding ANN-SNN feature map pair, as expressed in Equation (\ref{eq:stm_score}):
\begin{equation}
	\label{eq:stm_score}
	STM score = \frac{1}{T} \frac{1}{|\mathcal{C}^{t}|} \sum_{t=1}^{T} \sum_{\mathcal{C}^{t}} MSE (A_{s_{l}}^{t}, A_{a_{l}}).
\end{equation}
where $T$ represents the number of time steps, $|\mathcal{C}^{t}|$ denotes the number of candidate layer pairs, and $A_{s_{l}}^{t}$ and $A_{a_{l}}$ are the similarity matrices of ANN layer $_{a_{l}}$ and SNN layer $_{s_{l}}$, respectively. MSE measures the extent of semantic mismatches between the student SNN and the teacher ANN. A lower $STM score$ signifies fewer mismatched association semantics, equating to superior model performance. Practically, we calculate the $STM score$ (log-scale) values for each approach across training epochs and average them over the last 10 epochs, where they remain nearly unchanged.

Contrary to previous assumption, we find that existing ANN-to-SNN knowledge distillation methods either achieve very small improvements or result in degradation effects on STM scores, as shown in Table \ref{tab:mismatch_score1}. In other words, spatio-temporal mismatch of semantic information results in the loss of valuable knowledge during the knowledge distillation process. Our proposed approach diverges from the traditional paradigm by introducing self-attentive calibration. This innovative method aims to effectively transfer spatio-temporal semantic information by dynamically selecting suitable layer associations at each time step, departing from dependence on fixed teacher-student feature patterns.

\begin{table}[h]
	\centering
	\renewcommand{\arraystretch}{1}
		\setlength{\tabcolsep}{7pt}
		{\fontsize{10}{12}\selectfont
			\begin{tabular}{cccccc}
				\hline
				\multirow{2}{*}{\textbf{\scriptsize SNN}} & \multirow{2}{*}{\textbf{\scriptsize Dataset}} & \multirow{2}{*}{\textbf{\scriptsize Time Step}} & \multicolumn{3}{c}{\textbf{\scriptsize STM score ($\downarrow$)}}  \\ 
				& & &\scriptsize Baseline &\scriptsize KD  & \scriptsize FT   \\  \cline{4-6}
				\scriptsize VGG-11 &\scriptsize CIFAR-100 &\scriptsize 3 &\scriptsize 16.58 &\scriptsize 16.49  &\scriptsize 16.46 	 \\
				\scriptsize ResNet-18 &\scriptsize CIFAR-100 &\scriptsize 3 &\scriptsize 16.97 &\scriptsize 16.85  &\scriptsize 16.73 	\\	
				\scriptsize ResNet-18 &\scriptsize ImageNet &\scriptsize 4 &\scriptsize 22.81 &\scriptsize 22.97 &\scriptsize 22.68 	\\	\hline
		\end{tabular}}
		\begin{tablenotes}
		\footnotesize
		\item[1] Note: teacher ANNs for CIFAR-100 and ImageNet are ResNet-32x4 and ResNet-34, respectively. The symbol ($\downarrow$) indicates the smaller the better.
		\end{tablenotes}
		\caption{Evaluation of Spatio-Temporal Mismatch Score on CIFAR-100 and ImageNet}
		\label{tab:mismatch_score1}
	\end{table}
	
	\subsection{Formulation of Self-Attentive Calibration}
	In our methodology, each student layer at every time step seamlessly aligns itself with semantically matched target layers through attention allocation, as depicted in Figure. \ref{fig:overview} (a) and (b). The training process, guided by calibrated associations, prompts the student model to adeptly gather and integrate information from multiple layers at each time step, fostering a more tailored regularization. Furthermore, SASTC is versatile and can be applied in scenarios where the number of candidate layers differs between the teacher and student models. The ensemble of acquired feature patterns at time step $t$ in SASTC is denoted as $\mathcal{C}^{t} = \{(f^{t}_{s_{l}}, f_{a_{l}}) ~ | ~ \forall ~ f^{t}_{s_{l}} \in F^{t}_{s}, f_{a_{l}} \in F_{a}\}$, with the corresponding weight satisfying $\sum^{a_{L}}_{a_{l}=1} \eta^{t}_{(f^{t}_{s_{l}}, f_{a_{l}})}=1, \forall ~ f^{t}_{s_{l}} \in F^{t}_{s}$ at each time step. This weight $\eta^{t}_{(f^{t}_{s_{l}}, f_{a_{l}})} \in \mathbb{R}^{b \times T}$ signifies the degree to which the target layer $f_{a_{l}}$ is considered in the calibration of spatio-temporal semantic differences during ANN-to-SNN distillation. A more detailed exploration of these self-attentive weights will be provided subsequently. The feature maps from each time step of the student model are transformed into $a_{L}$ distinct forms, aligning with the spatial dimensions of each target layer for subsequent distance calculations, as indicated by 
	\begin{equation}
		\label{eq:align}
		\begin{split}
			\hat{f}^{t}_{s_{l}, a_{l}} = Proj ~ (f^{t}_{s_{l}} \in \mathbb{R}^{b \times c_{s_{l}} \times h_{s_{l}} \times w_{s_{l}}}, a_{l}), \\ f^{t}_{s_{l}} \in F^{t}_{s}, a_{l} \in [1, \dots, a_{L}],
		\end{split}
	\end{equation}
	with $\hat{f}^{t}_{s_{l}, a_{l}} \in \mathbb{R}^{b \times c_{a_{l}} \times h_{a_{l}} \times w_{a_{l}}}$. Each function $Proj(\cdot, \cdot)$ comprises a stack of two convolution layers with $3 \times 3$ and $1 \times 1$ to fulfill the requirement for an efficient transformation. This design choice is guided by previous research, which has illustrated the remarkable effectiveness of a $3 \times 3$ convolution layer and the pyramid architecture \cite{han2017deep}. 
	
	\subsubsection{Allocation with Self-attention}
	Previous research suggests that the abstraction of feature representations is closely associated with the layer depth \cite{Bengio2013Representation}. The semantic levels of these intermediates can vary between teacher and student architectures with differing capacities. Furthermore, we observe variations in the semantic level among the feature patterns of SNNs at different time steps, resulting in spatio-temporal information loss during distillation. To address these spatio-temporal differences in the student model and enhance the performance of feature transfer during distillation, each layer of the student model should be associated with the most semantically relevant target layer to derive its own regularization. Simple approaches such as random selection or forcing feature maps from the same layer depths to align may be inadequate due to the adverse effects resulting from semantic mismatched pairs.
	
	Inspired by the layer associations facilitated by attention mechanisms in ANNs \cite{vaswani2017attention}, we expand the self-attentive scheme from spatial calibration to spatio-temporal calibration. This extension presents a potentially viable solution for addressing the semantic mismatch problem and enhancing the overall distillation performance. Given that feature maps in SNNs, generated by similar instances, tend to cluster at distinct granularities across different time steps and layers, and similarly, in ANNs, these feature maps cluster based on their depth, the proximity of pairwise similarity matrices serves as a meaningful measure of inherent semantic similarity. These similarity matrices are computed as follows 
	\begin{equation}
		\label{eq:similarity_matrices}
		\begin{split}
			A^{t}_{s_{l	}} = R(f^{t}_{s_{l}}) \cdot R(f^{t}_{s_{l}})', ~ f^{t}_{s_{l}} \in F^{t}_{s},
			\\ A_{a_{l	}} = R(f_{a_{l}}) \cdot R(f_{a_{l}})', ~  f_{a_{l}} \in F_{a},
		\end{split}
	\end{equation}
	where $R(\cdot) : \mathbb{R}^{b \times c \times h \times w} \rightarrow \mathbb{R}^{b \times chw}$ represents a reshaping operation, and the symbol $'$ denotes the $transpose$ operation. Consequently, $A^{t}_{s_{l}}$ and $A_{a_{l}}$ yield $b \times b$ matrices.  More importantly, incorporating similarity matrices significantly mitigates the memory cost associated with large spatio-temporal dimensions ($T \cdot c_{s_{l} / a_{l}} \cdot h_{s_{l} / a_{l}} \cdot w_{s_{l} / a_{l}} \gg b$).
	
	Building upon the self-attention framework \cite{vaswani2017attention}, we independently project the pairwise similarity matrices of each student layer, originating from individual time steps, and each target layer, into two subspaces using a Multi-Layer Perceptron (MLP). This procedure endeavors to alleviate the influence of noise and sparsity, with the resulting vectors identified as $query$ and $key$. To expound further, for the $i$-th instance, the formulation can be articulated as follows: 
	\begin{equation}
		\label{eq:mlp}
		Q^{t}_{s_{l}}[i] = MLP_{Q}(A^{t}_{s_{l}}[i]), ~ K_{a_{l}}[i] = MLP_{K}(A_{a_{l}}[i]).
	\end{equation}			
	The parameters in $MLP_{Q}(\cdot)$ and $MLP_{K}(\cdot)$ are acquired through training to produce $query$ and $key$ vectors, shared across all instances. Subsequently, the calibrated weight $\eta^{t, i}_{(f^{t}_{s_{l}}, f_{a_{l}})}$ for the $i$-th instance is computed as follows:
	\begin{equation}
		\label{eq:cal_phi}
		\eta^{t, i}_{(f^{t}_{s_{l}}, f_{a_{l}})} = \frac{e^{Q^{t}_{s_{l}}[i]'  K_{a_{l}}[i]}}{\sum^{a_{L}}_{j=1} 	e^{Q^{t}_{s_{l}}[i]'  K_{j}[i]}}.
	\end{equation}
	
	The allocation based on attention offers a viable approach to alleviate the adverse effects stemming from spatio-temporal differences (mismatch in student-teacher layer pairs) and amalgamate beneficial guidance from multiple target layers. The complete training procedure, incorporating the proposed semantic calibration formulation, is succinctly outlined in Algorithm \ref{alg:training}.
	
	\subsubsection{Loss Function}
	In a mini-batch of size $b$, the student model generates multiple feature patterns spanning various time steps ($F^{t}_{s}, \dots, F^{T}_{s}$). Following dimensional projections and self-attentive calibration, we employ Mean-Square-Error (MSE) to align the raw pairwise similarity matrices of the teacher and student (referred to as the loss of SASTC),
	\begin{equation}
		\label{eq:loss_kd}
		\begin{split}
			\mathcal{L}_{SASTC} = \sum_{t} \sum_{l} \eta^{t}_{(f^{t}_{s_{l}}, f_{a_{l}})} Dist (f_{a_{l}}, Proj (f^{t}_{s_{l}}, 	a_{l})) 
			\\  = \sum^{b}_{i = 1} \sum^{T}_{t = 1} \sum^{a_{L}}_{a_{l} = 1} \sum^{s_{L}}_{s_{l} = 1} \eta^{t, i}_{(f^{t}_{s_{l}}, f_{a_{l}})} 	MSE(f_{a_{l}}[i], \hat{f}^{t}_{s_{l}, a_{l}}[i]),
		\end{split}
	\end{equation}
	as it demonstrated superior empirical performance.
	In this process, each feature map from the student model $f^{t}_{s_{l}}$ undergoes transformation via a projection function $Trans_{s} = Proj(\cdot, \cdot)$, while the target layers remain unchanged through identity transformation $Trans_{a}(\cdot) = I (\cdot)$. Multiplying the outcomes by the learned self-attentive distributions, the total loss is computed through a weighted summation of individual distances among feature maps from candidate teacher-student layer pairs at each time step. Consequently, the total loss of SASTC is expressed as:
	\begin{equation}
		\label{eq:loss_total}
		\mathcal{L}_{total} = \mathcal{L}_{KD} + \beta \mathcal{L}_{SASTC}. 
	\end{equation}
	\begin{equation}
		\label{eq:loss_soft}
		\mathcal{L}_{KD} = \mathcal{L}_{CE}(y_{i}, \sigma(O^{i}_{s})) + \alpha ^{2}\mathcal{L}_{KL}(\frac{\sigma(O^{i}_{a})}{\alpha}, \frac{\sigma(O^{i}_{s})}{\alpha}).
	\end{equation}
	where $\mathcal{L}_{CE}$ represents the standard cross-entropy loss (CE) between the predicted probabilities of the student model and the one-hot target, and $\mathcal{L}_{KL}$ denotes the KL divergence between $P^{i}_{s}$ and the soft targets of the teacher model $P^{i}_{a}$. 
	
	\subsection{Neuron Model and Surrogate Gradient}
	\subsubsection{LIF Neuron}
	We employ the LIF neuron model, which in discrete time, is described by:
	\begin{equation}
		\label{LIF_iter}
		\begin{split}
			u^{t} =\lambda u^{t-1}+ I^{t} , ~ o^{t}  = \Theta (u^{t} - V_{th}),
		\end{split}				
	\end{equation}
	where $u$ signifies the membrane potential, $I^{t}$ denotes pre-synaptic inputs, $\lambda (< 1)$ represents the constant leaky factor in the membrane potential, $V_{th}$ signifies the threshold membrane potential, $\Theta$ stands for the Heaviside step function, $o$ denotes the spike output propagating to the next layer, and the superscript $t$ indicates the time step. Following the emission of the spike output, the reset operation is delineated in 
	\begin{equation}
		u^{t} = u^{t} \cdot (1 - o^{t}).
	\end{equation}
	To minimize trainable parameters, all neurons share identical leak values $\lambda$ and threshold potentials $V_{th}$. For consistency across experiments, we set the initial membrane potential $u^{0}$ to $0$, the threshold $V_{th}$ to $1$, and the leaky factor $\lambda$ to $0.5$.
	
	\subsubsection{Triangle Shape Surrogate Gradient}
	In this study, prioritizing a balance between accuracy and computational efficiency, we choose the triangular surrogate gradient, as established in prior research \cite{deng2022temporal}. The mathematical expression for the triangular surrogate gradient is as follows:
	\begin{equation}
		\label{eq:gradient}
		\frac{\partial o}{\partial u} = \frac{1}{\gamma ^{2}} max(0, \gamma - |u - V_{th}|),
	\end{equation}
	with $\gamma$ set to $0.3$ based on previous works \cite{deng2022temporal}.
	
	\begin{algorithm}[tb]
		\caption{Self-Attentive Spatio-Temporal Calibration for ANN-to-SNN Knowledge Distillation}
		\label{alg:training}
		\textbf{Input}: 
		Training dataset $\mathcal{X} = {(x_{i}, y_{i})}^{n}_{i=1}$; A pre-trained ANN (teacher model) with parameter $\mathcal{W}^{a}$; An SNN (student model) with randomly initialized parameter $\mathcal{W}^{s}$;
		\begin{algorithmic}[1] 
			\WHILE{$\mathcal{W}^{s}$ is not converged}
			\STATE Sample a mini-batch $b$ size samples from $\mathcal{X}$ named $\mathcal{B}$.
			\STATE Obtain intermediate layers' presentations $\mathbf{F}^{t}_{s}$ across time steps and $\mathbf{F}_{a}$ by propagating $\mathcal{B}$ into $\mathcal{W}^{a}$ and $\mathcal{W}^{s}$. 
			\STATE Construct pairwise similarity matrices $A^{t}_{s_{l}}$ and $A_{a_{l}}$ as Equation (\ref{eq:similarity_matrices}). 
			\STATE Perform self-attention based spatio-temporal calibration as Equation (\ref{eq:mlp}-\ref{eq:cal_phi}).
			\STATE Spatially align feature patterns across time steps as Equation (\ref{eq:align}).
			\STATE Update parameters $\mathcal{W}^{s}$ by propagating backward the surrogate gradients as defined in Equation (\ref{eq:gradient}) of the loss in Equation (\ref{eq:loss_total}) and Equation (\ref{eq:loss_kd}). 
			\ENDWHILE
		\end{algorithmic}
	\end{algorithm}

	\section{Experiments}
	To demonstrate the effectiveness of our proposed self-attentive spatio-temporal calibration in ANN-to-SNN knowledge distillation, we conduct comprehensive experiments. We evaluate various ANN-SNN combinations using popular network architectures like VGG \cite{simonyan2014very}, ResNet \cite{he2016deep}, PyramidNet \cite{han2017deep}, and WRN \cite{zagoruyko2016wide} on static datasets. Meticulously designed experiments and ablation studies validate the effectiveness of SASTC in providing proper regularization for student models. We apply SASTC to neuromorphic datasets like DVS-Gesture and DVS-CIFAR10, demonstrating its robust generalization in noisy-label learning. Additionally, we offer a visual analysis of SASTC's success.
	
	Further analyses of temporal information dynamics are provided in Appendix 1, with demonstrations of robust generalization of few-shot learning in Appendix 2. Details on batch size, sensitivity, computational efficiency, running time, and memory consumption are summarized in Appendix 3. The experimental setup details are available in Appendix 4.
	
	\begin{table*}[ht]
		\footnotesize
		\centering
		\renewcommand{\arraystretch}{1}
			\setlength{\tabcolsep}{8pt}
			{\fontsize{10}{12}\selectfont
				\begin{tabular}{cccccccccccccc}
					\hline
					\textbf{\scriptsize SNN}  &  \multicolumn{3}{c}{\scriptsize Baseline}  &  ANN    &  \multicolumn{3}{c}{\scriptsize KD}  &    \multicolumn{3}{c}{\scriptsize Feature KD}  &\multicolumn{3}{c}{\textbf{\scriptsize SASTC}}  	\\	
					
					\textbf{(student)}  &\scriptsize T=2 &\scriptsize T=3 &\scriptsize T=7 &\scriptsize  (teacher) &\scriptsize T=2 &\scriptsize  T=3 &\scriptsize  T=7 &\scriptsize  T=2 &\scriptsize  T=3 &\scriptsize  T=7 &\scriptsize  T=2 &\scriptsize  T=3 &\scriptsize  T=7  \\	\hline
					\multicolumn{14}{c}{\textbf{\scriptsize CIFAR-10}}																																																											\\	\hline
					\multirow{3}{*}{\scriptsize VGG-11} & \multirow{3}{*}{\scriptsize 92.48} & \multirow{3}{*}{\scriptsize 92.80} & \multirow{3}{*}{\scriptsize 93.16} &\scriptsize ResNet-19 &\scriptsize 93.01 &\scriptsize 92.92 &\scriptsize 93.22 &\scriptsize 93.28 &\scriptsize 93.48 &\scriptsize 93.60 &\scriptsize 93.55 &\scriptsize 93.73 &\scriptsize 93.92      \\ 
					&  &  &  &\scriptsize Pyramidnet-20 &\scriptsize 92.98 &\scriptsize 93.07 &\scriptsize 93.52 &\scriptsize 93.04 &\scriptsize 93.28 &\scriptsize 93.36 &\scriptsize	93.52 &\scriptsize 93.70 &\scriptsize 93.64 	     \\
					&  &  &  &\scriptsize WRN-28-4 &\scriptsize 92.14 &\scriptsize 92.83 &\scriptsize 93.22 &\scriptsize 92.64 &\scriptsize 92.80 &\scriptsize 93.48 &\scriptsize 93.07 &\scriptsize 93.10 &\scriptsize 93.70 	\\		\cline{2-14}
					\multirow{2}{*}{\scriptsize ResNet-18}  & \multirow{2}{*}{\scriptsize 94.04} & \multirow{2}{*}{\scriptsize 94.12} & \multirow{2}{*}{\scriptsize 94.72} &\scriptsize WRN-28-4  &\scriptsize 94.24 &\scriptsize 94.44 &\scriptsize 94.68 &\scriptsize 94.92 &\scriptsize 95.44 &\scriptsize 95.76 &\scriptsize 95.12 &\scriptsize 95.24 &\scriptsize 95.48		     \\ 
					&  &  &  &\scriptsize Pyramidnet-20  &\scriptsize 94.16 &\scriptsize 94.24 &\scriptsize 94.48 &\scriptsize 94.60 &\scriptsize 95.16 &\scriptsize 95.48 &\scriptsize 94.92 &\scriptsize 95.00 &\scriptsize 95.16		\\	\cline{2-14}
					\multirow{3}{*}{\scriptsize WRN-16-2}  & \multirow{3}{*}{\scriptsize 89.40} & \multirow{3}{*}{\scriptsize 89.48} & \multirow{3}{*}{\scriptsize 90.80} &\scriptsize ResNet-19  &\scriptsize 92.36 &\scriptsize 92.52 &\scriptsize 93.36 &\scriptsize 92.16 &\scriptsize 93.13 &\scriptsize 93.26 &\scriptsize 94.12 &\scriptsize 94.16 &\scriptsize 94.12		     \\
					&  &  &  &\scriptsize Pyramidnet-20  &\scriptsize 92.04 &\scriptsize 92.60 &\scriptsize 93.28 &\scriptsize 92.86 &\scriptsize 93.18 &\scriptsize 93.51 &\scriptsize 93.16 &\scriptsize 93.76 &\scriptsize 93.96		     \\
					&  &  &  &\scriptsize WRN-28-4  &\scriptsize 91.72 &\scriptsize 92.20 &\scriptsize 92.92 &\scriptsize 92.16 &\scriptsize 93.08 &\scriptsize 93.44 &\scriptsize 93.16 &\scriptsize 93.28 &\scriptsize 93.72		     \\ \hline
					\multicolumn{14}{c}{\textbf{\scriptsize CIFAR-100}}																																																											\\	\hline
					\multirow{3}{*}{\scriptsize VGG-11}  & \multirow{3}{*}{\scriptsize 68.70} & \multirow{3}{*}{\scriptsize 69.76} & \multirow{3}{*}{\scriptsize 70.00} &\scriptsize VGG-13  &\scriptsize 73.44 &\scriptsize 74.52 &\scriptsize 75.24 &\scriptsize 74.01 &\scriptsize 74.43 &\scriptsize 75.11 &\scriptsize 74.60 &\scriptsize 74.88 &\scriptsize 76.36	     \\ 
					&  &  &  &\scriptsize ResNet-32x4  &\scriptsize 73.28 &\scriptsize 74.32 &\scriptsize 75.16 &\scriptsize 73.76 &\scriptsize 73.88 &\scriptsize 75.07 &\scriptsize 75.40 &\scriptsize 76.80 &\scriptsize 77.08		     \\
					&  &  &  &\scriptsize WRN-40-2  &\scriptsize 74.24 &\scriptsize 75.16 &\scriptsize 75.48 &\scriptsize 74.31 &\scriptsize 75.07 &\scriptsize 75.48 &\scriptsize 74.48 &\scriptsize 75.60 &\scriptsize 75.72 			    \\ \cline{2-14}
					\multirow{2}{*}{\scriptsize ResNet-18}  & \multirow{2}{*}{\scriptsize 70.56} & \multirow{2}{*}{\scriptsize 75.72} & \multirow{2}{*}{\scriptsize 76.40} &\scriptsize ResNet-32x4  &\scriptsize 77.68 &\scriptsize 78.00 &\scriptsize 78.64 &\scriptsize 77.01 &\scriptsize 77.96 &\scriptsize 78.09 &\scriptsize 80.28 &\scriptsize 80.24 &\scriptsize 80.68     		 \\
					&  &  &  &\scriptsize WRN-40-2  &\scriptsize 77.56 &\scriptsize 78.00 &\scriptsize 79.12 &\scriptsize 77.14 &\scriptsize 77.68 &\scriptsize 78.24 &\scriptsize 77.76 &\scriptsize 78.36 &\scriptsize 78.84		     \\ \hline
			\end{tabular}}
			\begin{tablenotes}
				\footnotesize
				\item[1] Note: baseline SNNs are trained from scratch using the same surrogate gradient as our distilled student models.
			\end{tablenotes}
			\caption{Top-1 Test Accuracy(\%) of Different ANN-to-SNN Distillation Approaches on CIFAR-10 and CIFAR-100 datasets}
			\label{tab:top1_cifar}
		\end{table*}	
		
		\subsection{Comparison to Conventional ANN-to-SNN Distillation Methods}	
		Top-1 test accuracy (\%) on CIFAR-10 and CIFAR-100 across seventeen distinct ANN-SNN combinations is illustrated in Table \ref{tab:top1_cifar}. Four of these combinations share similar architectures (WRN-16-2/28-4, VGG-11/13, ResNet-18/32x4), while the remaining nine are heterogeneous. Since the large memory consumption of traditional feature ANN-to-SNN KD method , we train the student SNN with different single layer combination settings and calculate their average value as the final result (\emph{i.e.,} Feature KD in Table \ref{tab:top1_cifar}).
		
		Table \ref{tab:top1_cifar} illustrates that SASTC demonstrates significant relative improvement across all compared methods on both CIFAR-10 and CIFAR-100 datasets, indicating its ability to effectively leverage intermediate information across time steps for superior distillation results. The most notable enhancements occur in the "WRN-16-2 \& ResNet-19" (T=2) and the "ResNet-18 \& ResNet-32x4" (T=2), with a 4.72\% improvement over baseline on CIFAR-10 and a 8.84\% improvement over baseline on CIFAR-100. Notably, when compared to the competitive distillation method feature KD, results with combinations of WRN-16-2/ResNet-19 on CIFAR-10 and all combinations on CIFAR-100 meet performance degradation phenomenon compared to the vanilla ANN-to-SNN KD method (the detailed explanation of these negative regularization effects is illustrated in the section of mechanism analysis). 
		
		\begin{table}[h]
			\centering
			\renewcommand{\arraystretch}{1}
			\setlength{\tabcolsep}{11pt}
			{\fontsize{10}{12}\selectfont
					\begin{tabular}{cccc}
						\hline
						\textbf{\scriptsize Method} &  \textbf{\scriptsize Architecture}  &  \textbf{\scriptsize Time Steps}  &  \textbf{\scriptsize Accuracy}	 \\	\hline
						\scriptsize Baseline                        &  \scriptsize ResNet-18         &        \scriptsize 4         &   \scriptsize  60.50            \\
						\scriptsize KD	                              &	  \scriptsize  ResNet-18         &		\scriptsize	4	      &	\scriptsize	61.37			  \\
						\scriptsize Feature KD                   &	 \scriptsize  ResNet-18         &		  \scriptsize  4       	 &   \scriptsize   61.01            \\
						\multirow{2}{*}{\textbf{\scriptsize SASTC}}           &	  \scriptsize ResNet-18         &		 \scriptsize  4         &		\scriptsize 62.52			  \\		
						&	 \scriptsize ResNet-34         &		  \scriptsize 4         &	\scriptsize	68.69			  \\		\hline
						\scriptsize Teacher (ANN)            &    \scriptsize ResNet-34         &       \scriptsize  1         &     \scriptsize  73.48            \\		\hline
				\end{tabular}}
			\caption{Top-1 Test Accuracy(\%)  of ANN-to-SNN Distillation Approaches on ImageNet dataset}
			\label{tab:top1_imagenet}
			\end{table}
			
			Moreover, we achieve significant improvements on ImageNet, which is illustrated in Table \ref{tab:top1_imagenet}. Despite training for only 90 epochs, SASTC improves ResNet-18 performance by 2.02\% over the baseline SNN. Additionally, SASTC achieves 1.15\% and 1.51\% improvements over vanilla and feature-based ANN-to-SNN KD on ResNet-18, respectively.  Notably, feature-based method shows significant negative regularization effects during the ANN-to-SNN distillation process on ImageNet.
			
			\subsection{Comparison to Existing SNN Training Methods}
			In this section, we present a comparison of our experimental results with previous conventional training methods, summarized in Table \ref{tab:comparison_static}. 
			
			On the CIFAR-10 dataset, our SASTC outperforms all existing approaches, achieving the highest accuracy and the lowest inference latency. Specifically, even with $T=2$, there is 0.62\% and 1.96\% increments compared to advanced TET \cite{deng2022temporal} and STBP-tdBN \cite{zheng2021going} with T=6, respectively. On the CIFAR-100 dataset, SASTC consistently demonstrates superior performance and faster inference, surpassing STBP-tdBN \cite{zheng2021going} by 9.56\% to 9.99\% and TET \cite{deng2022temporal} by 4.68\% to 5.96\% across all reported time steps. Notably, our method first outperforms the ANN counterpart with the spatio-temporal calibration on both CIFAR-10 and CIFAR-100, the relatively maximum increments are 0.51\% and 5.33\%, respectively.
			
			On the ImageNet dataset, the SASTC algorithm achieves a 3.90\% increment compared to TET \cite{deng2022temporal} with smaller 4 time steps. Although SEW-ResNet34 deviates from a typical SNN as it adopts the IF model and modifies the Residual structure, SASTC achieves 1.65\% improvement than SEW-ResNet \cite{fang2021incorporating}
			
			\begin{table}[h]
				\centering
				\renewcommand{\arraystretch}{1.1}
				\setlength{\tabcolsep}{0.01pt}
				{\fontsize{7}{7}\selectfont
						\begin{tabular}{ccccc}
							\hline
							\textbf{Dataset} & \textbf{Method} & \textbf{Architecture} & \textbf{Time Steps} & \textbf{Accuracy} \\ \hline
							\multirow{21}{*}{\rotatebox{90}{CIFAR-10}} &Norm \cite{Sengupta2019} &VGG-16 & 2500 & 91.55 \\
							&Norm \cite{han2020rmp} &VGG-16 & 2048 & 93.63 \\
							&Norm \cite{deng2021optimal} &VGG-16 & 16 & 92.29 \\
							&STBP	 \cite{wu2018spatio}                & CIFARNet & 12   & 89.83			\\
							&STBP NeuNorm \cite{wu2019direct} & CIFARNet & 12   & 90.53			\\
							&Hybrid \cite{rathi2020enabling}          &  ResNet-20 & 250 & 92.22			 \\
							&DIET-SNN \cite{rathi2021diet}        & ResNet-20	 & 10   & 92.54	          \\
							&TET \cite{deng2022temporal}           &  ResNet-19 & 6     & 94.50			\\
							&TSSL-BP \cite{zhang2020temporal} &	 CIFARNet	     & 5     &	91.41			 \\		\cline{2-5}
							&\multirow{3}{*}{STBP-tdBN \cite{zheng2021going}}   & \multirow{3}{*}{ResNet-19}	 & 6 & 93.16			\\  
							&                                                       &		                                            & 4 & 92.92			\\	
							&                                                       &		                                            & 2 & 92.34			\\	  \cline{2-5}
							& GLIF \cite{yao2022glif} & ResNet-19 & 2 & 94.44		\\
							& KDSNN \cite{xu2023constructing} & VGG-16 & 4 & 91.05		\\
							& Norm \cite{bu2023optimal} &	VGG-16	& 4	& 93.96		\\
							& TKS \cite{dong2024tks}	& ResNet-19	& 4	& 95.30		\\
							& SSCL-SNN \cite{zhang2024enhancing}	& ResNet-20	& 4	& 94.27		\\	\cline{2-5}
							&\multirow{3}{*}{\textbf{Ours}}	&  \multirow{3}{*}{ResNet-18} &	7  & 95.48       \\ 
							&                                                       &                                                   & 3 & 95.24         \\			
							&                                                       &                                                   & 2 & 95.12         \\		\cline{2-5}
							&ANN \cite{deng2022temporal}     & ResNet-19	  &  1		                &	94.97			\\	\hline
							\multirow{14}{*}{\rotatebox{90}{CIFAR-100}} &Hybrid \cite{rathi2020enabling} &  VGG-11	      &  125			    &	67.87			 \\
							&DIET-SNN \cite{rathi2021diet}                                 & VGG-16        & 5           & 69.67	          \\
							&TET \cite{deng2022temporal}       & ResNet-19     & 6	         & 74.72			\\  \cline{2-5}
							&\multirow{3}{*}{STBP-tdBN \cite{zheng2021going}}    & \multirow{3}{*}{ResNet-19} & 6		             &	71.12			\\	
							&                	                                   &		                                           & 4		             &	70.86			\\	
							&                                                      &		                                           & 2		             &	69.41			\\	\cline{2-5}
							& GLIF \cite{yao2022glif} & ResNet-19 & 2 & 75.48	\\
							& Norm \cite{bu2023optimal} &	VGG-16	& 4	& 69.62	\\
							& TKS \cite{dong2024tks}	& ResNet-19	& 4	& 76.20	\\
							& SSCL-SNN \cite{zhang2024enhancing}	& ResNet-19	& 2	& 78.79 	\\	\cline{2-5}
							&\multirow{3}{*}{\textbf{Ours}}	& \multirow{3}{*}{ResNet-18} & 7 & 80.68       \\	
							&                                                      &                                                 & 3  & 80.24      \\			
							&                                                      &                                                 & 2  & 80.28     \\	\cline{2-5}
							&ANN \cite{deng2022temporal}    & ResNet-19	  &  1		                & 75.35			\\	\hline
							\multirow{9}{*}{\rotatebox{90}{ImageNet}} & Hybrid \cite{rathi2020enabling} &  ResNet-34 & 250 & 61.48			 \\
							& SPIKE-NORM \cite{Sengupta2019} & ResNet-34 & 2500 & 69.96	          \\
							& STBP-tdBN	\cite{zheng2021going}     & Spiking-ResNet-34	 &  6	  &	   63.72			\\	
							& SEW ResNet \cite{fang2021incorporating}       & SEW-ResNet-34		&  4     &     67.04            \\
							& TET \cite{deng2022temporal}           &  Spiking-ResNet-34	   &  6	    &	64.79			\\ 
							& SSCL-SNN  \cite{zhang2024enhancing}           &  ResNet-34	   &  4	    &	66.78			\\ 
							& MS-ResNet \cite{hu2024advancing} & ResNet-18 & 6 & 63.10 \\ \cline{2-5}
							& \multirow{2}{*}{\textbf{Ours}} &  ResNet-18                   & 4  & 62.52 \\
							&                                                       &  ResNet-34                   &  4  & 68.69 \\ \hline
					\end{tabular}}
				\caption{Top-1 Test Accuracy(\%)  of Different SNN Methods on Static Datasets}
				\label{tab:comparison_static}
				\end{table}
				
				\subsection{Mechanism Analysis and Ablation Study}
				In this section, we delve into an experimental exploration of the negative regularization effect induced by manually specified layer associations across time steps. Furthermore, we provide evidence of the success of SASTC, supported by the proposed criterion and visual evidence.  
				
				\subsubsection{SASTC Improves Negative Regularization Effects}
				We conduct experiments on the CIFAR-10 dataset by training the student model exclusively with a specified teacher-student layer pair in various settings, and observe negative regularization effects that feature-pattern-based distillation with specific layer associations across time steps performs worse than vanilla ANN-to-SNN KD. The network architectures involved "VGG-11\& ResNet-19", "ResNet-18 \& Pyramidnet-20", and "WRN-16-2 \& WRN-28-4". The numbers of candidate target layers and student layers for each case are (4, 5), (3, 4), and (4, 4), respectively.
				
\begin{figure} [t]
					\centering
					\includegraphics[width=0.47\textwidth]{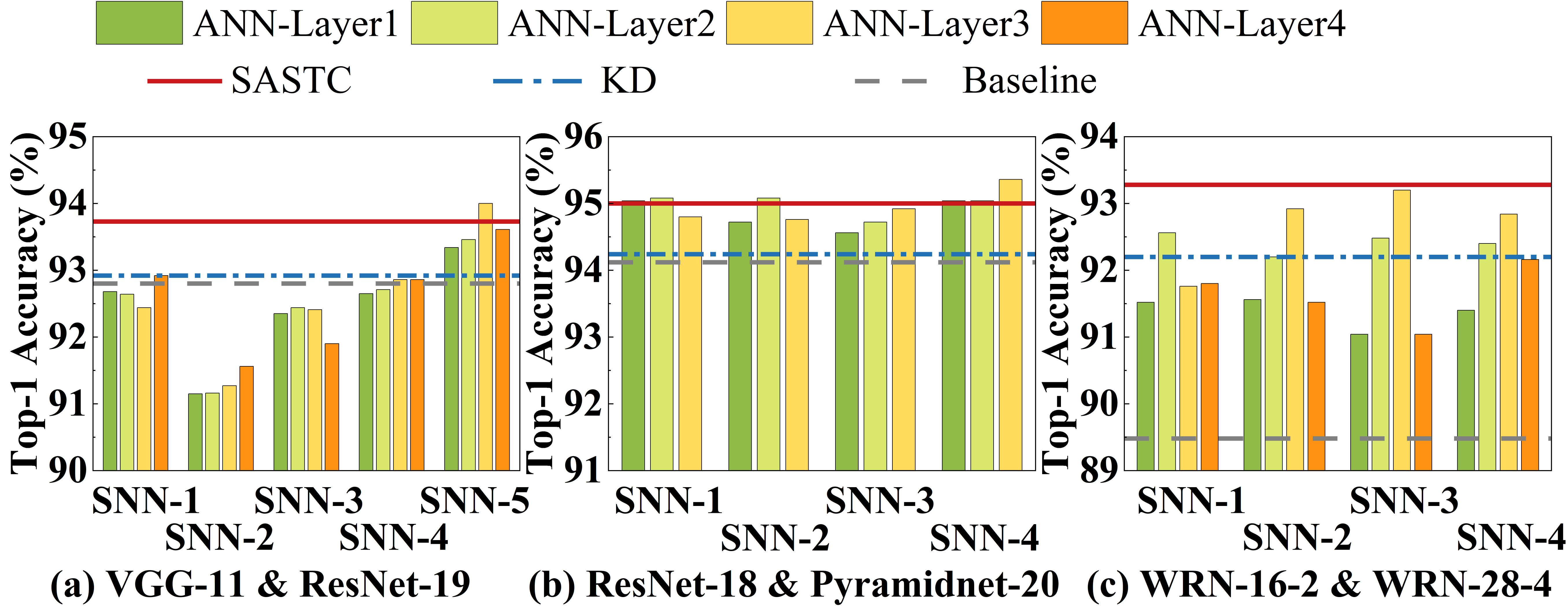}
					\caption{Illustration of negative regularization on CIFAR-10 with three model combinations. Each tick label of x-axis denotes an SNN (student) layer number. Different color bars indicate the results of different specified ANN-SNN layer combinations.}
					\label{fig:regulation_analysis}
				\end{figure}

				The outcomes of student models with 20, 12 and 16 ANN-SNN layer combinations under the three settings on CIFAR-10 are illustrated in Fig. \ref{fig:regulation_analysis}. Notably, all layer associations of SNN layer-2 and layer-3 in Fig. \ref{fig:regulation_analysis} (a) obtain extremely poor performance, likely due to highly sparse semantic information contained in layer-2 of VGG-11 student model. In addition, the performance of a student model significantly diminishes for certain layer associations across time steps (i.e., negative regularization effect), including SNN layer-1 to layer-4 in Fig. \ref{fig:regulation_analysis} (a) and (c), most like due to the substantial semantic mismatch. Notably, it is observed that the one-to-one layer matching scheme is non-optimal because better results can be obtained by leveraging information from a target layer with different depth, such as "SNN layer-5 \& ANN-layer3" in Fig. \ref{fig:regulation_analysis} (a), "SNN layer-4 \& ANN-layer3" in Fig. \ref{fig:regulation_analysis} (b) and "SNN layer-2 \& ANN-layer3" in Fig. \ref{fig:regulation_analysis} (c).
				
				Although training with specific hand-crafted layer associations may outperform SASTC in isolated cases like "SNN layer-5 \& ANN-layer3" in Fig. \ref{fig:regulation_analysis} (a) and "SNN layer-4 \& ANN-layer3" in Fig. \ref{fig:regulation_analysis} (b), SASTC consistently performs well across a large number of associations. It is particularly noteworthy considering that the knowledge of the best layer association for each network combination is not available in advance. Furthermore, instances where training with SASTC is inferior to the best layer association suggest potential refinements in our association strategy.
				
				\subsubsection{SASTC Achieves Semantic Matching during Knowledge Distillation}
				The results in Table \ref{tab:mismatch_score2} indicate that SASTC consistently attains the lowest spatio-temporal mismatch score throughout the training process compared to other approaches owing to our spatio-temporal calibration mechanism on both CIFAR-100 and ImageNet datasets. 
				
				Moreover, we provide evidence in Appendix 1 that SASTC optimizes the temporal information dynamics of SNNs, further illustrating the success and mechanism of our proposed method.
				\begin{table}[h]
					\centering
					\renewcommand{\arraystretch}{1}
					\setlength{\tabcolsep}{5pt}
					{\fontsize{10}{12}\selectfont
							\begin{tabular}{ccccccc}
								\hline
								\multirow{2}{*}{\textbf{\scriptsize SNN}} & \multirow{2}{*}{\textbf{\scriptsize Dataset}} & \multirow{2}{*}{\textbf{\scriptsize Time Step}} & \multicolumn{4}{c}{\textbf{\scriptsize STM score ($\downarrow$)}}  \\ 
								& & &\scriptsize Baseline &\scriptsize KD  &\scriptsize  FT & \textbf{\scriptsize SASTC}   \\  \cline{4-7}
								\scriptsize VGG-11 &\scriptsize CIFAR-100 &\scriptsize 3 &\scriptsize 16.58 &\scriptsize 16.49  &\scriptsize 16.46 &\scriptsize 16.30 	 \\
								\scriptsize ResNet-18 &\scriptsize CIFAR-100 &\scriptsize 3 &\scriptsize 16.97 &\scriptsize16.85 &\scriptsize 16.73 &\scriptsize 16.11 	\\	
								\scriptsize ResNet-18 &\scriptsize ImageNet &\scriptsize 4 &\scriptsize 22.81 &\scriptsize 22.97 &\scriptsize 22.68 &\scriptsize 22.09 	\\	\hline
						\end{tabular}}
						\begin{tablenotes}
							\footnotesize
							\item[1] Note: teacher ANNs for CIFAR-100 and ImageNet are ResNet-32x4 and ResNet-34, respectively. The symbol ($\downarrow$) indicates the smaller the better.
						\end{tablenotes}
					\caption{Evaluation of Spatio-Temporal Mismatch Score on CIFAR-100 and ImageNet}
					\label{tab:mismatch_score2}
					\end{table}
				
				\subsection{Extension, Application and Visual Analysis}
				\subsubsection{Extension to Neuromorphic Datasets}
				Previous works focus on utilizing complex architectures to tackle the challenging DVS-CIFAR10 task. However, these sophisticated models have failed to achieve satisfied performance and are susceptible to overfitting. Recently, a temporal efficient training approach has achieved the state-of-the-art accuracy with the streamlined VGGSNN architecture. In this paper, we conduct experiments on the DVS-Gesture and DVS-CIFAR10 datasets. As shown in Table \ref{tab:comparison_DVS}, our proposed method outperform the contemporary best-performing methods, the accuracy of SASTC increase to 97.92\% on DVS-Gesture and 83.60\% on DVS-CIFAR10 through the calibration of spatio-temporal semantic mismatches during ANN-to-SNN distillation. Consequently, our SASTC method significantly improves the processing temporal information ability of SNNs on neuromorphic tasks.
				
				\begin{table}[h]
					\centering
					\renewcommand{\arraystretch}{1}
					\setlength{\tabcolsep}{1pt}
					{\fontsize{10}{10}\selectfont
							\begin{tabular}{cccc}
								\hline
								\textbf{\scriptsize Method} &  \textbf{\scriptsize Architecture}  &  \textbf{\scriptsize Time Step}  &  \textbf{\scriptsize Accuracy}	   \\	\hline
								\multicolumn{4}{c}{\textbf{\scriptsize DVS-Gesture}}																	\\ \hline
								\scriptsize PLIF \cite{fang2021incorporating}                &	 \makecell[c]{\scriptsize c128k3s1-BN-PLIF- \\ \scriptsize MPk2s2*5-DPFC512- \\ \scriptsize PLIF-DP-FC110- \\ \scriptsize PLIF-APk10s10}	 & \scriptsize 20				   &\scriptsize 97.57			\\
								\scriptsize STBP-tdBN \cite{zheng2021going}            &\scriptsize CIFARNet	 & \scriptsize 40	&\scriptsize 96.87			\\
								\scriptsize SLAYER \cite{shrestha2018slayer}	      & \scriptsize 8 layers	   & \scriptsize 25  &\scriptsize 93.64			 \\
								\textbf{\scriptsize Ours}	&\scriptsize ResNet-18	     &\scriptsize 16              &\scriptsize 97.92               \\ \hline
								\multicolumn{4}{c}{\textbf{\scriptsize DVS-CIFAR10}}																	\\ \hline
								\scriptsize STBP-tdBN	\cite{zheng2021going}       & \scriptsize ResNet-19	      &\scriptsize  10			    &\scriptsize	67.80			 \\
								\scriptsize Streaming Rollout \cite{kugele2020efficient}      & \scriptsize DenseNet	        &\scriptsize 10			   &\scriptsize 66.80	          \\
								\scriptsize Conv3D \cite{wu2021liaf}   & \scriptsize LIAF-Net          &\scriptsize  10              &\scriptsize  71.70            \\
								\scriptsize LIAF \cite{wu2021liaf}          & \scriptsize LIAF-Net          &\scriptsize  10              &\scriptsize  70.40            \\ 
								\scriptsize TET \cite{deng2022temporal}             & \scriptsize VGGSNN          &\scriptsize10              &\scriptsize 83.17            \\ 
								\scriptsize SSCL-SNN  \cite{zhang2024enhancing}            & \scriptsize ResNet-19          &\scriptsize10              &\scriptsize 80.00	\\
								\textbf{\scriptsize Ours}	& \scriptsize VGGSNN	       &\scriptsize 10            &\scriptsize 83.60         \\		\hline
						\end{tabular}}
					\caption{Top-1 Test Accuracy(\%)  of Different SNN Methods on Neuromorphic Datasets}
					\label{tab:comparison_DVS}
					\end{table} 
					
					\subsubsection{Application to Robust Representation}
					In addition to evaluating the clean test set performance, we introduce noisy-label learning datasets by randomly perturbing 10\%, 20\%, 30\%, 40\%, and 50\% of labels in training images. As shown in Table \ref{tab:noisy-label}, training the lightweight SNN with SASTC extremely enhances its robustness compared to other ANN-to-SNN knowledge distillation approaches and the teacher ANN counterpart. 
					\begin{table}[h]
						\centering
						\renewcommand{\arraystretch}{1}
						\setlength{\tabcolsep}{4pt}
						{\fontsize{10}{12}\selectfont
								\begin{tabular}{ccccccc}
									\hline
									\makecell{\textbf{\scriptsize Percentage of} \\ \textbf{\scriptsize Noisy Labels}} & \textbf{\scriptsize 0\%} & \textbf{\scriptsize 10\%} & \textbf{\scriptsize 20\%} & \textbf{\scriptsize 30\%} & \textbf{\scriptsize 40\%}	& \textbf{\scriptsize 50\%}	\\	\hline
									\makecell{\scriptsize Teacher \\ (\scriptsize ResNet-32x4)}       &\scriptsize 79.42\% &\scriptsize 76.51\% &\scriptsize 73.63\% &\scriptsize 70.57\% &\scriptsize 68.08\% &\scriptsize 64.27\% 	\\	
									\scriptsize Baseline  &\scriptsize 69.76\% &\scriptsize 66.84\% &\scriptsize 64.00\% &\scriptsize 61.40\% &\scriptsize 59.84\%\scriptsize &\scriptsize 54.48\% 	\\	
									\scriptsize KD  &\scriptsize 74.32\% &\scriptsize 73.64\% &\scriptsize 72.64\% &\scriptsize 72.36\% &\scriptsize 71.36\% &\scriptsize 70.80\%		\\	
									\scriptsize Feature KD &\scriptsize 73.96\% &\scriptsize 63.68\% &\scriptsize 63.36\% &\scriptsize 63.24\% &\scriptsize 62.96\% &\scriptsize 61.84\%			 \\	
									\textbf{\scriptsize SASTC}  & \textbf{\scriptsize 77.08\%} & \textbf{\scriptsize 75.49\%}& \textbf{\scriptsize 74.77\%} & \textbf{\scriptsize 74.62\%} & \textbf{\scriptsize 74.32\%} &\textbf{\scriptsize 73.87\%} \\		\hline
							\end{tabular}}
							\begin{tablenotes}
								\footnotesize
								\item[1] Note: SNNs adopt three time steps.
							\end{tablenotes}
						\caption{Noisy-Label Learning: Top-1 Test Accuracy(\%)  of "VGG-11 \& ResNet-32x4" Combination on CIFAR-100}
						\label{tab:noisy-label}
						\end{table}
					
					\subsubsection{Visualization Analysis of SASTC}
\begin{figure} [h]
						\centering
						\includegraphics[width=0.47\textwidth]{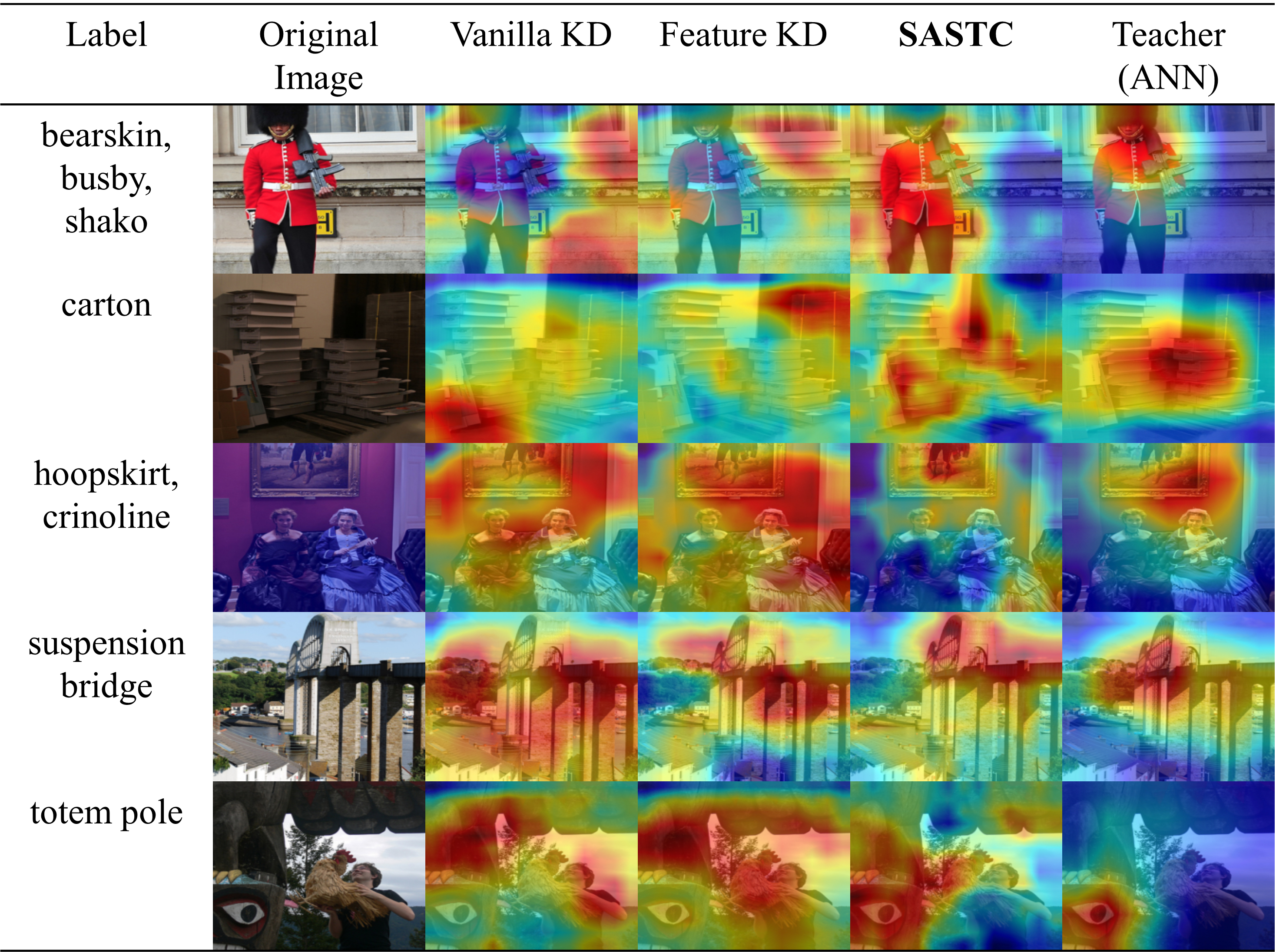}
						\caption{Spike Activation Map (SAM) visualization of ANN-to-SNN distillation approaches on ImageNet. The red regions highlight areas deemed important for model inference.}
						\label{fig:visualize}
					\end{figure}

					To visually elucidate the advantages of SASTC, we randomly selected several images from ImageNet, and highlight regions deemed crucial for predicting the respective labels by utilizing Spike Activation Map (SAM) \cite{kim2021visual}. As depicted in Figure \ref{fig:visualize}, SASTC consistently centralizes class-discriminative regions and excels in capturing more semantically related information, resembling the teacher model, while the compared methods scatter them in the surroundings.
					
\section{Conclusion}
This study focuses on mitigating performance degradation due to spatio-temporal semantic mismatches and negative regularization in conventional ANN-to-SNN knowledge distillation methods. We propose a self-attentive mechanism to learn layer association weights across different time steps, enabling semantically aligned knowledge transfer. Qualitative and quantitative evidence validate SASTC's spatio-temporal calibration capability. Extensive experiments demonstrate that SASTC consistently outperforms various SNN training approaches and distillation schemes. SASTC also shows strong generalization across tasks and network architectures, excelling in robust representation. 

\section{Acknowledgments}
This work was supported in part by the National Natural Science Foundation of China under Grants 62336007, in part by the Key R\&D Program of Zhejiang under Grant 2022C03011, in part by the Starry Night Science Fund of Zhejiang University Shanghai Institute for Advanced Study under Grant SN-ZJU-SIAS-002, and in part by the Fundamental Research Funds for the Central Universities.

\bibliography{aaai25}		

\clearpage



\section*{Supplementary material}

# Appendix 1

## SASTC Optimizes Temporal Information Dynamics of SNN

To quantify the Fisher information and further analysis its distribution of different ANN-to-SNN distillation methods, we introduce the empirical Fisher Information Matrix trace, denoted as $FIM^t$ in Equation (1):

$$FIM^t = \frac{1}{n} \sum_{j=1}^{n} ||\nabla_{\mathcal{W}} log P_{\mathcal{W}}(y|x_{k\leq t}^n)||^2, \quad (1)$$

where $k$ denotes the index of time step, and $P_{\mathcal{W}}(y|x)$ denotes approximate posterior distribution of a network with weight parameters $\mathcal{W}$, input instance $x$ from data distribution $\mathcal{X}$ and output $y$.

In Fig. 1, we visualize temporal Fisher information of three different ANN-to-SNN distillation methods using the "VGG-11 & ResNet-19" combination on the CIFAR-10 dataset, along with the baseline as a reference. The Fisher information decreases notably with ANN-to-SNN distillation, and with training progression, the Fisher information with SASTC decreases more rapidly towards the final stable state compared to other distillation methods. In terms of temporal information distribution, ANN-to-SNN distillation can effectively supervise the optimization of the weight parameters, and our proposed SASTC is more effective than other ANN-to-SNN distillation methods.

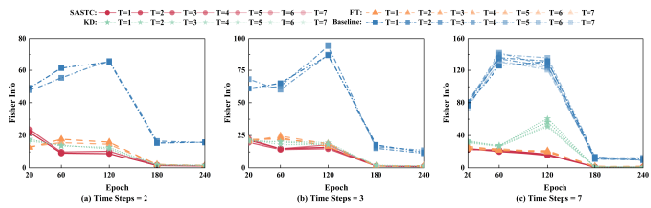

Figure 1: Illustration of temporal dynamics across training epochs of "VGG-11 & ResNet-19" combination on CIFAR-10. Four distinct colors represent three different distillation methods and the baseline model. Various levels of transparency within each color signify different time steps.

## Application to Few-Shot Learning

Furthermore, we assess the performance of SASTC in few-shot learning and noisy-label learning scenarios. These experiments consistently demonstrate that SASTC effectively leverages training data and exhibits relative robustness to noisy perturbations.

We randomly select 25%, 50%, and 75% of training images from each class in CIFAR-100 to create several few-shot learning datasets. As depicted in Table 1, SASTC consistently outperforms the compared ANN-to-SNN distillation approaches and conventional training methods in all settings. The improvement becomes more significant as the number of available training images decreases. Notably, SASTC surpasses the teacher ANN counterpart and when trained with only 25% and 50% of the original dataset, which further prove the advantages of our SASTC.

Table 1: Few-Shot Learning: Top-1 Test Accuracy(%) of "VGG-11 & ResNet-32x4" Combination on CIFAR-100

| Percentage of Training Images | 25% | 50% | 75% | 100% |
|---|---|---|---|---|
| Teacher (ResNet-32x4) | 60.97% | 70.87% | 75.04% | 79.42% |
| Baseline | 46.48% | 60.32% | 66.76% | 69.76% |
| KD | 56.41% | 66.94% | 71.62% | 74.32% |
| Feature KD | 50.14% | 57.97% | 61.87% | 73.96% |
| **SASTC** | **66.19%** | **72.19%** | **74.44%** | **77.08%** |

Note: SNNs adopt three time steps.

## Additional Analysis

### Batch Size Analysis

We conduct experiments to analysis the influence of choosing different batch size in our proposed method, since our calculation of similarity matrices and self-attentive calibration is based on batch-level. To consider eliminating the impact of other factors, we introduce the vanilla ANN-to-SNN KD method for comparison as well as directly trained SNNs as the baseline. Results are summarized in Table 2. Although batch size is a crucial hyperparameter that influences training performance with the SGD optimizer (Keskar et al. 2016), the SNN obtained through SASTC consistently outperforms other SNNs in all cases.

Table 2: Impact of the Batch Size: Top-1 Test Accuracy(%) of "VGG-11 & ResNet-32x4" Combination on CIFAR-100

| Batch Size | 16 | 32 | 64 | 128 | 256 |
|---|---|---|---|---|---|
| Baseline | 72.96 | 73.52 | 73.48 | 73.24 | 72.44 |
| KD | 72.96 | 73.52 | 74.60 | 74.32 | 73.60 |
| SASTC | **74.56** | **75.80** | **76.4** | **76.36** | **76.24** |

Note: SNNs adopt three time steps.

### Sensitivity Analysis

We evaluate the impact of the hyperparameter beta on the performance of SASTC across CIFAR-10 and CIFAR-100 datasets. Combinations of "VGG-11 & ResNet-19" with 3 time steps and "VGG-11 & ResNet-32x4" with 3 time steps are adopted on CIFAR-10 and CIFAR-100 respectively. Comparisons are made between SASTC with varying beta values, the vanilla ANN-to-SNN knowledge distillation approach (KD), and baseline SNNs. The hyperparameter $\beta$ for SASTC ranges from 100 to 1500 at intervals of 100, while it remains 0 for KD, resulting in two horizontal lines in Fig. 2.

On the CIFAR-100 dataset, except for the $\beta$ setting of 100, SASTC consistently outperforms both KD and the baseline in all other cases. Particularly, at the optimal hyperparameter setting ($\beta$ = 700), SASTC exceeds KD and the baseline by 1.62 and 6.18 absolute accuracy, respectively. Similarly, on CIFAR-10, Fig. 2 demonstrates that SASTC consistently outperforms KD and the baseline across different $\beta$ settings. At the optimal beta setting of 1300, our method surpasses KD and the baseline by 1.12 and 1.24 absolute accuracy, respectively. These findings demonstrate the effectiveness of our proposed method across a wide range of the $\beta$ hyperparameter search space.

### Comparison of Computation Efficiency

For computation energy analysis, we count the total spiking operations (SOPs) to estimate the energy consumption of different distillation SNNs compared with their ANN counterparts (Akopyan et al. 2015). Specifically, we adopt "VGG-11 & ResNet-32x4" combination, then sample 1024 samples and calculate the mean value of SOPs. Early stopping is also used to alleviate overfitting with a tolerance of 20 epochs. We also introduce the direct trained SNN (baseline) for comparison. In conservative estimation, SNNs consume 0.9pJ energy per accumulation (AC) and 4.6pJ energy per multiplication and accumulation (MAC) as measured in 45nm CMOS technology (Rathi and Roy 2021). However, further study proves that the energy consumption of SNNs on specified hardware design can be reduced by 12× to 77 fJ/SOP (Rathi and Roy 2021). SNNs in this paper have multiplication operations in the first layer, while the other layers only have addition operations.

Table 3 summaries the results of various SNN training methods with different time steps. Notably, the SNN obtained through SASTC has 79.40% Top-1 test accuracy and 614.1 uJ energy consumption with 2 time steps, which achieves both the higher performance and higher efficiency compared to directly trained (baseline), vanilla distillation (KD) and feature distillation (feature KD) SNNs with 7 time steps. In addition, the energy is significantly efficient when MAC degrades to AC. For instance, our SASTC trained SNNs outperform the ANN counterpart of 0.18%($T = 2$), 1.02%($T = 3$) and 1.46%($T = 7$) with only 4.91%, 6.73% and 15.71% energy consumption respectively.

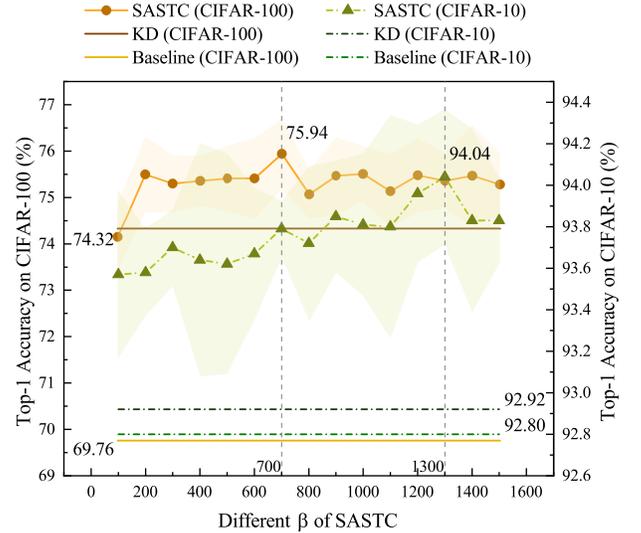

Figure 2: Impact of the hyperparameter $\beta$.

Table 3: Comparions of Energy Consumption under Different Time Steps of "ResNet-18 & ResNet-32x4" Combination on CIFAR-100

| Method | Time Steps | T=2 | T=3 | T=7 |
|---|---|---|---|---|
| Baseline | Accuracy (%) | 70.56 | 75.72 | 76.40 |
| | Total Energy (uJ) | 388.3 | 571.1 | 1303.9 |
| KD | Accuracy (%) | 77.68 | 78.00 | 78.64 |
| | Total Energy (uJ) | 373.7 | 560.4 | 1271.5 |
| Feature KD | Accuracy (%) | 77.2 | 78.16 | 78.40 |
| | Total Energy (uJ) | 378.3 | 566.3 | 1259.3 |
| SASTC | Accuracy (%) | **79.40** | **80.24** | **80.68** |
| | Total Energy (uJ) | 614.1 | 841.5 | 1965.6 |
| ANN | Accuracy (%) | | 79.22 | |
| | Total Energy (uJ) | | 12511.2 | |

Note: SNNs adopt three time steps.

### Comparison of Running Time and Memory Consumption

We analysis the time and memory cost with the "VGG-11 & ResNet-32x4" combination on CIFAR-100. The overall running time of vanilla ANN-to-SNN KD, feature-based ANN-to-SNN KD and SASTC is about 3.6, 3.0, 2.8 h with 2 time steps, 4.0, 3.2, 3.9 h with 3 time steps, and 6.0, 4.3, 3.6 h

with 7 time steps, respectively. The overall GPU memory usage of these methods is about 0.8 to 1.2 GB based on our training settings and computing infrastructure in Appendix 2. A. Our proposed SASTC obtain similar performance in shorter training time (SASTC with T=2 VS. KD and Feature KD with T=7). While the training time and memory usage vary across methods and ANN-SNN combinations, the comparison results above underscore the superiority of SASTC. Additionally, SASTC requires more computational resources, including time and GPU memory, to achieve its superior performance, which represents an acceptable trade-off.

## Experimental Setup

### Datasets

- **Static Datasets** focus on popular image classification tasks, including CIFAR-10 (Krizhevsky, Hinton et al. 2009), CIFAR-100 (Krizhevsky, Hinton et al. 2009), and ImageNet (Krizhevsky, Sutskever, and Hinton 2012). Data augmentation is applied on CIFAR-10 and CIFAR-100 by padding each training image with 4 pixels on each side and extracting a randomly sampled $32 \times 32$ crop or its horizontal flip. The ImageNet dataset includes over 1.2 million training images, 50,000 validation images, and 100,000 test images across 1,000 classes. For ImageNet, each training image is randomly sampled as a $224 \times 224$ crop or its horizontal flip without any padding operation.

- **Neuromorphic Datasets** consist of DVS-Gesture (Amir et al. 2017) and DVS-CIFAR10 (Li et al. 2017) datasets, capturing events reflecting changes in pixel brightness. In the DVS-Gesture dataset, the task involves classifying 2-dimensional event-stream inputs into 11 hand gesture classes. The input event bins are scaled to a $64 \times 64$ size before being fed into the model. DVS-CIFAR10 provides each label with 9,000 training samples and 1,000 test samples.

### Training Details

We employ the stochastic gradient descent (SGD) optimizer with a momentum of 0.9 in all experiments. For CIFAR-10 and CIFAR-100, the initial learning rate is set to 0.01, and models are trained for 240 epochs with the learning rate divided by 10 at 150, 180, and 210 epochs. The mini-batch size is fixed at 128, and the weight decay is set to $5 \times 10^{-4}$. Results are reported as means (standard deviations) over 3 trials. Regarding ImageNet, the initial learning rate is 0.1, reduced by a factor of 10 at 30 and 60 out of the total 90 training epochs. The mini-batch size is set to 256, and the weight decay is $1 \times 10^4$, with results reported from a single trial. For consistency and fair comparison, we maintain a temperature $\alpha$ of 4 for the ANN-to-SNN KD loss. For static datasets, the hyper-parameter $\beta$ of SASTC is set to 1300 on CIFAR-10, 700 on CIFAR-100 and 50 on ImageNet, respectively. For neuromorphic datasets, models are trained using the SGD optimizer for 300 epochs. The initial learning rate is set to 0.1 and divided by 10 at epochs 150, 180, 210, and 260. Additionally, set hyperparameter $\beta$ to 0.01, 0.1, and 1 at the first, second, and last 100 epochs, respectively. In practice, the building blocks (i.e., *student layers* and *target layers*) are notably fewer than the total layer count in both SNN and ANN.

### Computing Infrastructure

All experiments are conducted using PyTorch (Paszke et al. 2019) on a server equipped with four NVIDIA RTX 3090 GPUs, and another server with four NVIDIA RTX A800 GPUs.

# Appendix 2

Table 1: Training and Inference Times of Different ANN-to-SNN Distillation Approaches on ImageNet.

| Method | Architecture | Time Steps | Training Time (h) | Inference Time (mins) |
|---|---|---|---|---|
| Baseline | ResNet-18 | 4 | 1.34 | 1.01 |
| KD | ResNet-18 | 4 | 1.36 | 1.01 |
| Feature | ResNet-18 | 4 | 1.97 | 1.12 |
| SASTC (Ours) | ResNet-18 | 4 | 2.50 | 1.07 |
| | ResNet-34 | 4 | 4.01 | 1.93 |

Table 2: Training and Inference Times per Epoch of Different ANN-to-SNN Distillation Approaches on CIFAR datasets

| (dent) | Baseline T=2 Train | Infer | T=3 Train | Infer | T=7 Train | Infer | ANN (teacher) | KD T=2 Train | Infer | T=3 Train | Infer | T=7 Train | Infer | Feature KD T=2 Train | Infer | T=3 Train | Infer | T=7 Train | Infer | SASTC (Ours) T=2 Train | Infer | T=3 Train | Infer | T=7 Train | Infer |
|---|---|---|---|---|---|---|---|---|---|---|---|---|---|---|---|---|---|---|---|---|---|---|---|---|---|
| | | | | | | | CIFAR-10 | | | | | | | | | | | | | | | | | | |
| VGG-11 | 36.77 | 12.11 | 37.70 | 12.46 | 48.17 | 12.64 | ResNet-19 | 43.50 | 19.18 | 46.50 | 19.57 | 62.16 | 19.84 | 74.21 | 10.18 | 81.42 | 10.49 | 120.49 | 13.88 | 99.78 | 13.24 | 102.24 | 13.78 | 117.35 | 14.99 |
| | | | | | | | Pyramidnet-20 | 41.98 | 19.60 | 45.11 | 19.48 | 60.80 | 19.57 | 73.73 | 10.14 | 80.26 | 10.54 | 135.68 | 13.07 | 99.23 | 13.37 | 103.36 | 12.54 | 123.66 | 15.40 |
| | | | | | | | WRN-28-4 | 41.96 | 19.57 | 45.84 | 19.11 | 62.29 | 19.71 | 97.36 | 11.56 | 109.23 | 11.70 | 125.93 | 18.12 | 100.93 | 14.60 | 102.50 | 13.82 | 131.22 | 16.01 |
| ResNet-18 | 51.94 | 12.58 | 66.34 | 12.89 | 71.26 | 13.02 | WRN-28-4 | 53.26 | 13.16 | 68.63 | 12.69 | 84.36 | 12.31 | 90.95 | 12.62 | 105.11 | 12.40 | 190.38 | 13.12 | 122.67 | 13.22 | 161.36 | 13.03 | 221.31 | 12.17 |
| | | | | | | | Pyramidnet-20 | 54.35 | 12.11 | 67.99 | 12.76 | 86.51 | 12.91 | 85.68 | 14.02 | 90.58 | 12.99 | 185.36 | 13.12 | 122.87 | 13.02 | 160.93 | 13.47 | 224.01 | 11.89 |
| WRN-16-2 | 84.99 | 12.81 | 95.33 | 12.79 | 112.36 | 13.87 | ResNet-19 | 100.93 | 15.22 | 136.91 | 14.98 | 376.60 | 15.80 | 114.22 | 13.78 | 156.11 | 13.06 | 356.33 | 16.08 | 171.01 | 15.12 | 241.43 | 15.07 | 311.81 | 15.07 |
| | | | | | | | Pyramidnet-20 | 94.75 | 14.90 | 132.36 | 14.78 | 344.37 | 15.28 | 107.43 | 12.29 | 152.44 | 13.11 | 339.51 | 16.89 | 166.14 | 15.04 | 233.89 | 15.32 | 301.67 | 15.58 |
| | | | | | | | WRN-28-4 | 95.04 | 14.86 | 134.35 | 15.04 | 359.11 | 14.86 | 118.75 | 12.78 | 151.44 | 13.71 | 328.38 | 16.13 | 177.25 | 15.06 | 250.11 | 15.13 | 306.22 | 15.29 |
| | | | | | | | CIFAR-100 | | | | | | | | | | | | | | | | | | |
| VGG-11 | 69.21 | 12.71 | 73.25 | 13.12 | 90.11 | 13.41 | VGG-13 | 86.77 | 13.61 | 90.19 | 17.44 | 153.80 | 18.83 | 101.22 | 12.33 | 125.41 | 12.91 | 215.12 | 12.76 | 129.99 | 13.01 | 130.35 | 13.53 | 185.73 | 13.28 |
| | | | | | | | ResNet-32x4 | 90.49 | 13.89 | 103.85 | 18.03 | 157.44 | 17.69 | 108.24 | 12.75 | 127.25 | 12.12 | 172.49 | 12.14 | 108.09 | 18.41 | 125.58 | 13.67 | 164.58 | 13.69 |
| | | | | | | | WRN-40-2 | 88.67 | 18.46 | 103.43 | 18.45 | 157.44 | 18.69 | 100.89 | 12.41 | 124.88 | 12.76 | 170.62 | 13.01 | 121.88 | 12.90 | 125.64 | 12.74 | 156.59 | 13.04 |
| ResNet-18 | 66.82 | 12.29 | 77.21 | 13.92 | 93.52 | 13.87 | ResNet-32x4 | 88.22 | 13.58 | 98.57 | 18.47 | 259.07 | 18.90 | 91.74 | 14.06 | 103.78 | 14.52 | 145.46 | 14.83 | 114.85 | 20.03 | 157.89 | 20.24 | 192.65 | 18.48 |
| | | | | | | | WRN-40-2 | 87.99 | 13.06 | 99.79 | 17.00 | 350.14 | 17.99 | 93.12 | 14.01 | 101.29 | 14.17 | 136.13 | 15.84 | 115.23 | 19.65 | 157.81 | 20.32 | 181.89 | 18.99 |

					
											
\end{document}